\documentclass{article}

\usepackage{ml4cps}

\usepackage[utf8]{inputenc} 
\usepackage[T1]{fontenc}    
\usepackage{amsmath}        
\usepackage{amssymb}
\usepackage{mathtools}
\usepackage[hidelinks,hypertexnames=false]{hyperref} 
\usepackage{url}            
\usepackage{xurl}           
\usepackage{booktabs}       
\usepackage{amsfonts}       
\usepackage{nicefrac}       
\usepackage{microtype}      
\usepackage{cleveref}       
\usepackage{graphicx}
\usepackage[numbers]{natbib} 
\usepackage{doi}

\usepackage{bm}
\usepackage{xcolor}
\usepackage{colortbl}
\usepackage{amsthm}
\usepackage{pgfplots}
\pgfplotsset{compat=1.18}
\usepackage{float}   
\usepackage{etoolbox}

\apptocmd{\thebibliography}{\raggedright}{}{}

\newtheorem{assumption}{Assumption}

\crefname{proposition}{Proposition}{Propositions}
\Crefname{proposition}{Proposition}{Propositions}
\crefname{remark}{Remark}{Remarks}
\Crefname{remark}{Remark}{Remarks}
\crefname{assumption}{Assumption}{Assumptions}
\Crefname{assumption}{Assumption}{Assumptions}

\newcommand{\T}{^{\mathsf{T}}}        
\newcommand{\E}{\mathbb{E}}           
\newcommand{\R}{\mathbb{R}}           
\newcommand{\mass}{m_b}               
\newcommand{\kb}{\bar{k}_b}           
\newcommand{\Sigb}{\Sigma_b}          
\newcommand{\vcb}{v^{c}_{b}}          
\newcommand{\score}{\mathrm{score}}   
\newcommand{\MX}{M_X}                 
\newcommand{\KX}{K_X}                 

\newcommand{\ostar}{o^{\star}}        
\newcommand{\ohat}{\hat{o}}           

\title{\textbf{COBS}: \textbf{C}umulant \textbf{O}rder \textbf{B}lock \textbf{S}parse Attention}

\date{}

\newif\ifuniqueAffiliation
\uniqueAffiliationtrue

\ifuniqueAffiliation 
\author{
	Alexander Tian\thanks{Equal contribution.}
	\quad Aditya Ghai\footnotemark[1]
	\quad Sanjit Neelam
	\quad Zaal Vasania
	\quad Akshay Mishra\\
	MatX\\
	\texttt{\{alexander,adi,sanjit,zaal,akshay\}@matx.com}
}
\else
\usepackage{authblk}

\newbox{\orcid}\sbox{\orcid}{\includegraphics[scale=0.06]{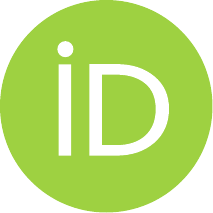}} 
\author[1]{%
	\href{https://orcid.org/0000-0000-0000-0000}{\usebox{\orcid}\hspace{1mm}First author}%
}
\author[1,2]{%
	\href{https://orcid.org/0000-0000-0000-0000}{\usebox{\orcid}\hspace{1mm}Second author}%
}
\affil[1]{Affiliation, Address}
\affil[2]{Affiliation, Address}
\fi

\renewcommand{\shorttitle}{COBS: Cumulant Order Block Sparse Attention}

\hypersetup{
pdftitle={COBS: Cumulant Order Block Sparse Attention},
pdfsubject={Second-order block selection for sparse attention},
pdfauthor={Alexander Tian, Aditya Ghai, Sanjit Neelam, Zaal Vasania, Akshay Mishra},
pdfkeywords={sparse attention, long-context, block selection, cumulant expansion, KV cache},
}

\begin{document}

\maketitle

\begin{abstract}
Block sparse attention is a hardware friendly way to alleviate the key--value (KV) cache read bottleneck in large language models (LLMs). However, it is not prevalent among leading open-weight LLMs, which rely instead on dense attention or fine-grained selection, thereby motivating our analysis. We study DeepSeek's Native Sparse Attention (NSA)~\cite{nsa} as a representative method, whose three-branch design lets us isolate block selection, the most challenging and consequential stage. We formalize selection and reduce it to ranking blocks by a single quantity, the \emph{attention mass}: the sum of a block's attention scores. We show that if selection retrieves the blocks with the largest attention mass, block sparse attention can match the quality of dense attention. However, computing the exact attention mass requires reading every key, so the problem of block selection ultimately reduces to approximating this mass from a compact summary instead of the full keys. Via a cumulant expansion, we show why existing methods falter: their selection strategies attempt to estimate the attention mass, but are confined to a first-order approximation. Therefore, we propose \textbf{COBS} (Cumulant Order Block Sparse Attention), an attention method that builds on NSA, incorporating a novel selector that stores a compressed second-order statistic per block. On the 32k RULER long-context retrieval benchmark~\cite{ruler}, COBS raises the NSA baseline's mean score from $0.2999$ to $0.8195$, approaching dense attention at $0.9040$ and closing about $86\%$ of the gap, while using only $1.21\times$ the KV cache read traffic of the NSA baseline and $15.15\times$ less read traffic than dense. The same model preserves short-context behavior and attains lower position-wise negative log-likelihood (NLL) than dense attention in our comparison.

\end{abstract}

\section{Introduction}
\label{sec:introduction}

Transformer inference at long context is bottlenecked by reading the key--value (KV) cache: at each decode
step, attention reads the keys and values of every past token, so decoding is limited by memory
bandwidth rather than compute and leaves hardware underutilized. Sparse attention reduces this cost by reading the keys and values of only a fraction of past tokens.

\begin{figure}[t]
  \centering
  \includegraphics[width=0.92\linewidth]{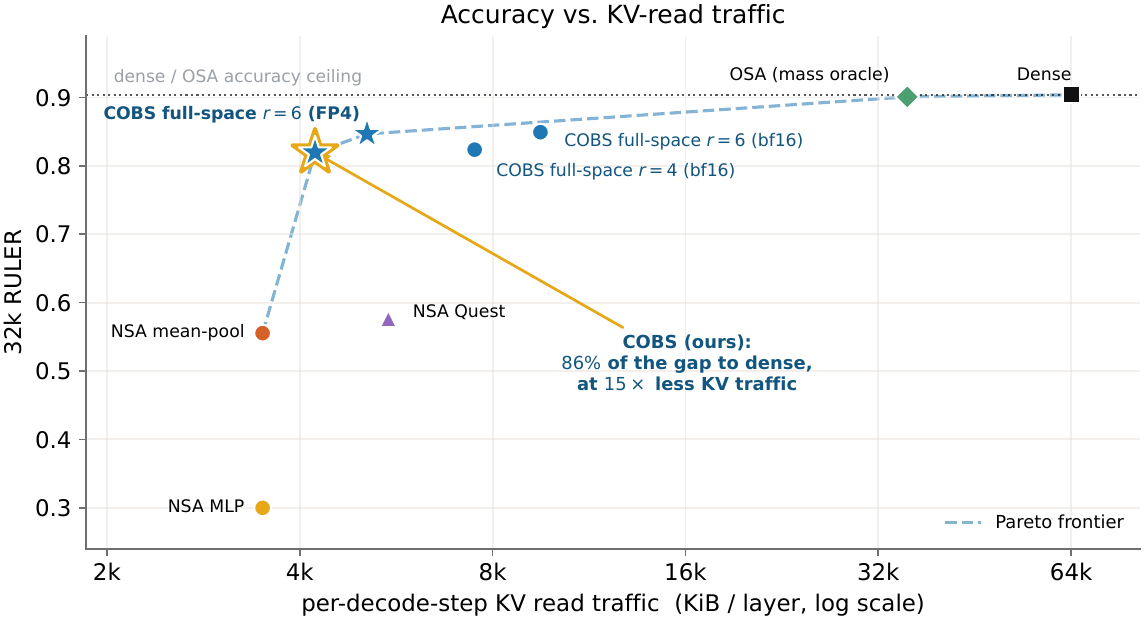}
  \caption{\textbf{COBS approaches dense accuracy at a fraction of the KV cache read traffic.}
  Accuracy on 32k RULER versus per-decode-step KV cache read traffic (KiB per layer, log scale; see
  \Cref{tab:bandwidth}). The plotted FP4 COBS configurations lie on the Pareto frontier (dashed); the
  highlighted star is COBS (adaptive $s{\approx}85$, $r{=}4$, FP4), which recovers most of the gap
  between dense and NSA MLP at $1.21\times$ NSA MLP's KV cache read traffic. NSA MLP and NSA Quest are
  dominated; NSA mean-pool anchors the low-traffic end of the frontier, and OSA buys its accuracy by
  re-reading all keys.}
  \label{fig:pareto}
\end{figure}

Sparse attention methods differ in what they keep and how they choose it. Fixed-pattern methods attend
to predetermined local windows and global tokens (Longformer~\cite{longformer}, BigBird~\cite{bigbird});
KV-eviction methods drop low-importance tokens (StreamingLLM~\cite{streamingllm}, H2O~\cite{h2o},
SnapKV~\cite{snapkv}); and low-rank methods compress the keys themselves~\cite{loki}. Closest to our
setting are \emph{query-aware} block selectors, which summarize contiguous blocks of the cache offline
and, per query, run fine-grained attention over the highest-scoring blocks; this family includes
Quest~\cite{quest} and the learned block selector of Native Sparse Attention (NSA)~\cite{nsa}. We
return to these methods in \Cref{sec:related_work}.

Among these families, block sparsity is arguably the most hardware friendly: attending to contiguous
blocks yields contiguous memory accesses, and a coarse block granularity requires a far smaller
top-$k$ than selecting individual tokens. Despite this affinity for hardware, block sparsity is largely
absent from leading open-weight LLMs. Recent releases instead span the alternatives: dense
grouped-query attention (GQA~\cite{gqa}; e.g.\ Hunyuan Hy3~\cite{hunyuanhy3}), latent KV compression (Multi-head
Latent Attention (MLA)~\cite{deepseekv2}, used by Kimi K2.6~\cite{kimik26}), local windows interleaved with periodic global layers (Gemma 4~\cite{gemma4}), and
fine-grained token selection (DeepSeek Sparse Attention (DSA)~\cite{deepseekv32}, adopted in modified
form by GLM-5.2~\cite{glm52}).
Even the closest case, DeepSeek-V4's CSA~\cite{deepseekv4}, selects over blocks only four tokens wide
and pairs them with a compression branch (HCA) that performs no selection at all. Large-block
selection, the form most amenable to hardware, is thus conspicuously underused, raising the question of
what limits block sparse methods and whether the gap to dense attention can be closed.

We study this question through NSA, a representative query-aware block sparse method. NSA pairs a
coarse stream with a top-$k$ block \emph{selection} branch, so selection alone decides which
distant blocks are seen at full resolution while its scoring summary stays cacheable, cleanly
isolating the mechanism we analyze (\Cref{sec:background}). Our findings target NSA but extend to the
broader family of \emph{first-order} block selectors, which score each block from a cached,
query-independent summary.

With selection isolated, the question becomes what a cacheable selector should compute. Under explicit
assumptions, block selection reduces to ranking blocks by their \emph{attention mass}
(\Cref{sec:selection_oracle}), and an oracle (OSA) ranking by the exact mass essentially matches dense
attention, so the obstacle is estimating the mass rather than the criterion. Existing selectors estimate it
only to first order and miss how relevance curves with the query direction, whereas our proposed method, COBS, restores the
omitted second-order term from a compressed key covariance cached per block. We
evaluate and ablate COBS at long context, where it closes most of the gap between first-order selection
and dense attention.

\paragraph{Contributions.}
\begin{itemize}
  \item We formalize block selection as preserving the true attention outputs and, under explicit
  assumptions, reduce it to ranking blocks by their attention mass, with an oracle (OSA) confirming
  this criterion essentially recovers dense attention performance (\Cref{sec:selection_oracle}).
  \item We give a cumulant-expansion view of block selection (\Cref{sec:cumulant}) that explains the
  limitation of first-order selectors: they capture the first-order term but discard the within-block
  covariance and higher-order terms needed to approximate the oracle.
  \item We propose \textbf{COBS} (Cumulant Order Block Sparse Attention, \Cref{sec:method}), an NSA-based sparse attention method whose block
  selector is derived from the cumulant expansion, storing the within-block covariance in compressed
  form per block (a low-rank factorization, a query-subspace projection, and FP4 quantization) at a
  small multiple of the cost of mean-pool scoring.
  \item We evaluate on our 11-task 32k RULER configuration (\Cref{sec:experiments}), where COBS raises
  mean score from $0.2999$ (NSA baseline) to $0.8195$, approaching full (dense) attention at $0.9040$
  and closing about $86\%$ of the gap, while using only $1.21\times$ the KV cache read traffic of the
  NSA baseline and $15.15\times$ less read traffic than dense attention (\Cref{fig:pareto}). Ablations
  isolate the main design choices.
\end{itemize}

\section{Background}
\label{sec:background}

\paragraph{Notation.}
At a decode step a query $q\in\R^D$, with $D$ the per-head dimension, attends over a KV cache of past
tokens with keys $k_r,v_r\in\R^D$. Dense attention forms scores $s_r$, weights $\propto e^{s_r}$, and outputs
the weighted average of values. We absorb the usual attention scale into $q$. We partition the cache into contiguous \emph{blocks} $b$, each
holding $L$ tokens, and write the block \emph{mass} $\mass$ and value centroid $\vcb$ as
\begin{equation}
  s_r=q\T k_r,\qquad \mass=\sum_{r\in b}e^{s_r},\qquad
  \vcb=\frac{1}{\mass}\sum_{r\in b}e^{s_r} v_r.
  \label{eq:bg-defs}
\end{equation}
We use multi-head attention (MHA)~\cite{attention} and grouped-query
attention (GQA)~\cite{gqa}; under GQA a group of $G$ query heads shares one KV head, so the
model has $H$ KV heads in total, and quantities indexed by $g$ range over the heads in a group.

\paragraph{NSA and its three branches.}
NSA~\cite{nsa} composes three attention branches whose outputs are gated and summed. (i) A
\emph{compression} branch attends over a coarse stream of pooled block representations. (ii) A
\emph{selection} branch ranks blocks by a lightweight per-block score, keeps the top-$k$, and
runs fine-grained attention over the kept blocks. (iii) A \emph{sliding-window} branch covers
recent local context.

The selection branch is the focus of this paper: it alone determines which
blocks fine-grained attention ever sees, so its fidelity bounds the quality of the whole method, and
its scoring summary is what must be cached.

\paragraph{What the selection branch must do.}
Selection must identify, before reading a block's full keys and values, the blocks that
contribute most to the attention output for the current query. NSA scores a block by an inner product between the query and a cached summary vector produced by a small MLP over the block's keys; although this pooling is nonlinear in the keys,
the summary enters the score only to first order in the query~\cite{nsa}. \Cref{sec:selection_oracle} derives
the attention mass surrogate used as our oracle, and \Cref{sec:related_work} shows how the
first-order score relates to that surrogate.

\paragraph{The cacheability constraint.}
A selector is \emph{cacheable} if its per-block summary can be precomputed and stored, then
reused across decode steps without re-reading the block. This forces the summary to be
\emph{independent of the decode query} $q$; for instance, any function of the block's keys alone satisfies this. Every design choice in this paper respects this constraint (we never store a
query-dependent quantity per block), and it is precisely this constraint that makes the
\emph{cumulant order} of the stored summary the binding limitation (\Cref{sec:cumulant}).

\section{The Selection Oracle}
\label{sec:selection_oracle}

The selection oracle formalizes block selection as reconstructing dense per-head attention outputs
from a top-$k$ block subset. We derive the exact single-head error from dropping blocks, then a value-agnostic reconstruction
bound whose minimizer gives the mass criterion. Index the $H$ KV heads by $h$ and the $G$ query
heads sharing a KV head by $g$ (\Cref{sec:background}); per-head quantities carry superscript
$(g,h)$. The selector chooses one block set $S^{(h)}$ per KV head, shared by its $G$ query heads
(we write $S$ when the head is clear from context).

\subsection{Dense head output and exact single-head error}
\label{subsec:oracle-mixture}

Within a head, normalizing the block masses $\mass$ (\Cref{eq:bg-defs}) by the full softmax
denominator $Z=\sum_i e^{s_i}$ gives block probabilities $P_b=\mass/Z$, so the dense head output
is a $P_b$-weighted mixture of the block centroids $\vcb$. Then, a selector that keeps a block set $S$
(dropping its complement $S^c$ with dropped mass $\tau=\sum_{b\in S^c}P_b$)
renormalizes over the kept blocks:
\begin{equation}
  \ostar=\sum_b P_b\,\vcb ,\qquad
  \ohat_S=\frac{\sum_{b\in S}P_b\,\vcb}{1-\tau}.
  \label{eq:oracle-true}
\end{equation}
Subtracting the second from the first and collecting terms gives the
exact per-head reconstruction error,
\begin{equation}
  \ostar-\ohat_S=\frac{1}{1-\tau}\sum_{b\in S^c}P_b\big(\vcb-\ostar\big),
  \label{eq:oracle-exact}
\end{equation}
which is the renormalized sum over dropped blocks of probability mass times
the deviation of each centroid from the true output.

\subsection{The per-KV-head selection objective}
\label{subsec:oracle-residual}

We define the oracle directly at the head-output level and optimize each KV head independently.
For a fixed KV head $h$, the shared top-$k$ block set controls the sum of the group's head-output
reconstruction errors:
\begin{equation}
  \mathcal{E}^{(h)}(S)=\sum_{g=1}^{G}\big\lVert(\ostar-\ohat_S)^{(g,h)}\big\rVert .
  \label{eq:oracle-multihead}
\end{equation}
The assumptions below turn this objective into a value-agnostic bound to minimize.

\subsection{Assumptions}
\label{subsec:oracle-assumptions}

\begin{assumption}[A1: Value-agnosticism]\label{ass:value-blind}
Within each head the centroid deviations share a common magnitude bound across blocks,
$\big\lVert(\vcb-\ostar)^{(g,h)}\big\rVert\le c^{(g,h)}$. We adopt this relaxation as the
deviations are nontrivial to estimate, since they require the exact head output $\ostar$ which we do not have.
\end{assumption}

\begin{assumption}[A2: Disregard $c$]\label{ass:gamma}
The per-head coefficients $c^{(g,h)}$ (the deviation-magnitude bounds of
\Cref{ass:value-blind}) are ignored for ranking. We also adopt this relaxation as these bounds are unknown.
\end{assumption}

\begin{assumption}[A3: Linear relaxation]\label{ass:linear}
For $G>1$ the per-head penalty $\sum_g\tau^{(g)}/(1-\tau^{(g)})$ is replaced by its leading-order
(linear) term $\sum_g\tau^{(g)}$, rendering the per-block score additive.
\end{assumption}

\subsection{The GQA selection score}
\label{subsec:oracle-relax}

Under \Cref{ass:value-blind,ass:gamma,ass:linear}, minimizing the per-KV-head objective
\eqref{eq:oracle-multihead} reduces to ranking blocks by an additive per-block score
(derived in \Cref{app:selection-derivation}), maximized by the top-$k$ blocks under
\begin{equation}
  \boxed{\ \score_b^{(h)}=\sum_{g=1}^{G}\frac{m_b^{(g,h)}}{Z^{(g,h)}},\qquad
    Z^{(g,h)}=\sum_{b'}m_{b'}^{(g,h)}\ }
  \label{eq:prop-gqa}
\end{equation}
The relaxation is valid when each dropped mass $\tau^{(g)}$ is small, so ranking by
\eqref{eq:prop-gqa} is a reasonable oracle rather than an exact optimum. For multi-head attention
($G=1$), minimizing the bound is exactly keeping the top-$k$ blocks by mass $\mass$, equivalently by
log-mass $\ln\mass$, so \Cref{ass:gamma,ass:linear} are unnecessary in this case (\Cref{app:selection-derivation}).

\noindent Crucially,~\eqref{eq:prop-gqa} gives the value-agnostic mass criterion that DeepSeek's
NSA~\cite{nsa} approximates with a cached summary. Our contribution is therefore a justification of the
mass criterion and a better \emph{estimator} of it (\Cref{sec:cumulant,sec:method}).

\subsection{Oracle Sparse Attention (OSA): a mass-oracle reference}
\label{subsec:osa}

The analysis above establishes the mass criterion as the target of selection, but leaves open how well
that criterion performs, and how much of the gap to dense
attention it can close. \textbf{OSA} (Oracle Sparse Attention) answers this by applying the selection mass criterion with the \emph{exact} masses, removing any error a
cached summary would introduce. Concretely, OSA extends NSA with only a single change. It computes the true block masses $m_b^{(g,h)}$ by reading the full block keys and ranks blocks
by the normalized GQA score of~\eqref{eq:prop-gqa}, leaving every other component of NSA intact. Because OSA reads every block's keys to form the exact masses, it provides sparsity only on value reads while key traffic stays dense. Its exact-mass ranking is therefore impractical to deploy, and we use OSA purely as a diagnostic.

On our 11-task 32k RULER
configuration (\Cref{sec:experiments}), OSA essentially matches dense attention, reaching a mean score
of $0.9010$ against dense's $0.9040$ and closing $99.5\%$ of the gap from the NSA baseline, which remains far below. Selecting the right blocks, when done with the
true masses, thus recovers nearly all of dense attention's long-context accuracy. What separates
existing block sparse methods from this ceiling is therefore not the selection criterion but rather the
\emph{estimation} of the mass from a cacheable summary.
\section{The Cumulant Expansion}
\label{sec:cumulant}

\Cref{sec:selection_oracle} showed that block selection reduces to ranking blocks by their mass
$\mass$, and that the difficulty lies in estimating $\mass$ from a compact, query-independent summary. To expose
what such a summary can capture, we expand the mass, and its logarithm, in the statistics of the
block's keys.

Write the mass as $L$ times an expectation over the block's empirical key distribution $X$:
\begin{equation}
  \mass=L\cdot\frac1L\sum_{r\in b}e^{s_r}=L\,\E_r\!\big[e^{q\T k_r}\big]=L\,\MX(q),
  \qquad \MX(q)=\E_X\!\big[e^{q\T X}\big],
  \label{eq:cgf-mgf}
\end{equation}
so $\MX$ is the moment generating function of the keys and
\begin{equation}
  \ln\mass=\ln L+\KX(q),\qquad \KX(q)=\ln\MX(q)
  \label{eq:cgf-kgf}
\end{equation}
$\KX$ is its cumulant generating function. The standard multivariate cumulant expansion, equivalently the
Taylor expansion of $\KX$ around $q=0$~\cite[Ch.~2]{cgf}, is
\begin{equation}
  \KX(q)=q\T\kappa_1+\tfrac12\,q\T\kappa_2\,q+\tfrac16\sum_{ijk}(\kappa_3)_{ijk}q_iq_jq_k+\cdots,
  \label{eq:cgf-expansion}
\end{equation}
with first two cumulants the block mean key and the within-block key covariance (derived in \Cref{app:cumulants}),
\begin{equation}
  \kappa_1=\kb=\frac1L\sum_{r\in b}k_r,\qquad
  \kappa_2=\Sigb=\frac1L\sum_{r\in b}(k_r-\kb)(k_r-\kb)\T .
  \label{eq:cgf-cumulants}
\end{equation}

Equivalently, $\mass=L\,e^{\KX(q)}$: a block's mass is fixed entirely by the cumulants of its keys.
Our method works from this form and its low-order truncations, keeping $\kb$ and $\Sigb$ as the
per-block summary (\Cref{sec:method}).

\section{Method}
\label{sec:method}

We present our method \textbf{COBS}, Cumulant Order Block Sparse Attention, as a sequence of additive improvements over a single controlled NSA
baseline (the exact baseline configuration is fixed in \Cref{sec:experiments}). The headline
change is \Cref{subsec:secondorder}: storing a compressed within-block covariance per block. The remaining
subsections are the estimation and storage optimizations that reduce its KV cache read traffic at fixed
selection quality.

\subsection{NoPE in the compression and selection branches}
\label{subsec:nope}
We remove rotary position encoding (RoPE) from the compression and selection branches, a
no-position-encoding (NoPE) scheme, so block summaries depend only on content. With RoPE, keys at different positions are rotated by different
angles, so pooling over a block mixes content with position and injects positional spread into the
summary. NoPE strips out this compression noise, letting each block summary focus on content;
precise relative position is retained only where it is needed, in the local window. We report NoPE
as a standalone additive change that on its own empirically improves NSA (\Cref{sec:experiments}).

\subsection{Second-order truncation: store \texorpdfstring{$\kb$}{the mean key} and a compressed \texorpdfstring{$\Sigb$}{covariance}}
\label{subsec:secondorder}
Truncating the cumulant expansion~\eqref{eq:cgf-expansion} at second order gives the core estimator:
\begin{equation}
  \boxed{\ \ln\mass\approx \ln L+q\T\kb+\tfrac12\,q\T\Sigb\,q\ }
  \label{eq:trunc}
\end{equation}
Note that the \emph{first-order} term is simply a mean-pool over the block keys:
\begin{equation}
  \hat\ell_b=q\T\kb,
  \label{eq:firstorder}
\end{equation}
while the second-order term $\tfrac12\,q\T\Sigb\,q$ is the curvature that mean-pooling omits (\Cref{fig:concept}).

Under GQA the
second-order mass estimate enters the group score of \Cref{eq:prop-gqa} through
\begin{equation}
  \begin{aligned}
    \widehat{\score}_b^{(h)}&=\sum_{g=1}^{G}\frac{1}{\widehat Z^{(g,h)}}
      \exp\!\Big(q^{(g,h)\top}\bar k_b^{(h)}+\tfrac12\,q^{(g,h)\top}\Sigma_b^{(h)}\,q^{(g,h)}\Big),\\
    \widehat Z^{(g,h)}&=\sum_{b'}\exp\!\Big(q^{(g,h)\top}\bar k_{b'}^{(h)}+\tfrac12\,q^{(g,h)\top}\Sigma_{b'}^{(h)}\,q^{(g,h)}\Big).
  \end{aligned}
  \label{eq:gqa-secondorder}
\end{equation}
Here $\widehat Z^{(g,h)}$ is the second-order estimate of the normalizer $Z^{(g,h)}$ of~\eqref{eq:prop-gqa} after the common block-length factor $L$ in $\mass=L\,e^{\KX(q)}$ is cancelled from numerator and denominator.
We therefore store, per block, the mean key $\kb$ and a compressed form of the covariance $\Sigb$,
and score blocks with this second-order estimate. The covariance summary is used for block \emph{selection} only: COBS retains a mean-pool compression branch, so selection and compression use different summaries. We find that even the first-order mean-pool approximation
already improves on DeepSeek's NSA MLP selection summary (\Cref{subsec:headline}).

\subsection{Covariance compression}
\label{subsec:compress}
For $D > L$, the $O(D^2)$ covariance costs more memory
than the $LD$ keys it summarizes, thus requiring compression of $\Sigb$. We use a low-rank
spectral decomposition keeping the top $r$ eigendirections of $\Sigb$,
\begin{equation}
  \Sigb\approx\sum_{i=1}^{r}\lambda_i u_i u_i\T=\sum_{i=1}^{r} \xi^{\mathrm{fs}}_i (\xi^{\mathrm{fs}}_i)\T,\qquad \xi^{\mathrm{fs}}_i=\sqrt{\lambda_i}\,u_i,
  \label{eq:lowrank}
\end{equation}
where the standard spectral decomposition is rewritten by folding $\sqrt{\lambda_i}$ into each
eigenvector.
We store the block mean $\kb$ ($D$ floats) and the $r$ scaled eigenvectors $\xi^{\mathrm{fs}}_i$ ($rD$
floats), where the superscript fs denotes full-space, for $D+rD$ floats per block.

\subsection{Subspace method}
\label{subsec:subspace}
The rank-$r$ approximation stores covariance directions in the full key space $\R^D$, but ranking depends on $\Sigb$ only through the scalar quadratic form $q\T\Sigb q$. Let $U_Q\in\R^{D\times s}$ hold the top $s$ eigenvectors of the query second moment $\E[q q\T]$, spanning the $s$-dimensional query subspace. We keep the covariance only within this subspace,
\begin{equation}
  B_b=U_Q\T\Sigb\,U_Q\in\R^{s\times s},
  \label{eq:subspace}
\end{equation}
and, projecting the query as $\tilde q=U_Q\T q\in\R^s$ with $\Pi=U_Q U_Q\T$, score with the projected quadratic form
\begin{equation}
  q\T\Sigb\,q\ \approx\ q\T\,\Pi\,\Sigb\,\Pi\,q\ =\ \tilde q\T B_b\,\tilde q ,
  \label{eq:subspace-quad}
\end{equation}
which is exact when $\Pi q=q$; the adaptive $s$ below keeps the out-of-subspace residual $\lVert(I-\Pi)q\rVert$ negligible. Rather than store the $s\times s$ matrix $B_b$, we take a rank-$r$ spectral decomposition in the projected space,
\begin{equation}
  B_b\approx\sum_{i=1}^{r}\xi^{\mathrm{ss}}_i(\xi^{\mathrm{ss}}_i)\T,\qquad \xi^{\mathrm{ss}}_i\in\R^s,
  \label{eq:subspace-lowrank}
\end{equation}
where the $\xi^{\mathrm{ss}}_i$ (ss denoting subspace) are the stored scaled eigenvectors of $B_b$. Subspace COBS stores the
$\xi^{\mathrm{ss}}_i$ and full-space ablations store the $\xi^{\mathrm{fs}}_i$ of~\eqref{eq:lowrank}; the
estimation, storage, and quantization below apply identically to either, so we write $\xi_i$ for the
stored factors wherever the distinction is immaterial. We set $s$ per layer as a function of the rank capturing $90\%$ of the query spectral energy, averaged over heads. This preserves near-full 32k
RULER selection quality while reducing the stored eigenvector cost from $rD$ to $rs$ floats per
block. At $r=4$, this is $340$ rather than $512$ floats, a ${\approx}1.5\times$ reduction. The subspace descriptor is thus an inference-time compression of the full-space one: starting from the corresponding trained full-space checkpoint, we leave the model weights fixed and replace each block's covariance descriptor by its query-subspace projection.

\subsection{Quantization}
\label{subsec:quant}
We quantize the stored scaled eigenvectors $\xi_i$ to the E2M1 FP4 format, keeping one
fp32 scale per eigenvector and the block mean at bf16 precision. On the rank-4 descriptor this is
essentially lossless: 32k RULER changes by only $+0.0013$ (bf16 $0.8238$ to FP4 $0.8251$), while
shrinking the covariance-factor descriptor ${\approx}3.8\times$ (from $1024$ to $272$ bytes per block,
excluding the bf16 mean) on top of the low-rank and subspace reductions of
\Cref{subsec:compress,subsec:subspace}; additional quantization results are reported in \Cref{subsec:quant-ablation}.

\subsection{Cost and KV cache read traffic}
\label{subsec:cost}
\paragraph{Per-decode-step scoring.} We describe the subspace descriptor here; the full-space variant is identical after dropping the projection and using $q$ in place of
$\tilde q$. At each decode step we form the projected query $\tilde q=U_Q\T q$ once, then evaluate the
curvature term from the stored subspace factors,
\begin{equation}
  q\T\Sigb\,q\approx\sum_{i=1}^{r}(\xi_i\T\tilde q)^2 ,
  \label{eq:gram}
\end{equation}
i.e. $r$ inner products in the $s$-dimensional subspace, costing $O(rs)$ per block (plus one
$O(sD)$ query projection shared across all blocks) against $O(D)$ for
mean-pool.

\paragraph{Eigenvector computation (Gram trick).} The scaled eigenvectors $\xi_i$ are recomputed
once per block (once every ${\approx}L$ decode steps) as the block fills. Let $\tilde K\in\R^{L\times D}$
be the centered key matrix with rows $(k_r-\kb)\T$, so $\Sigb=\tfrac1L\tilde K\T\tilde K$. Rather
than forming the full $D\times D$ covariance, we apply the Gram trick to the $L\times L$ Gram matrix
$\tfrac1L\tilde K\tilde K\T$: a unit eigenpair $(\lambda_i,w_i)$ maps to the unit eigenvector
$u_i=\tilde K\T w_i/\sqrt{L\lambda_i}$ of $\Sigb$ at the same eigenvalue $\lambda_i$, giving the
stored $\xi_i=\sqrt{\lambda_i}\,u_i=\tilde K\T w_i/\sqrt{L}$. For $D>L$ this avoids materializing the
full covariance: forming the Gram matrix costs $O(L^2D)$, the small eigendecomposition costs
$O(L^3)$, and recovering the top $r$ scaled eigenvectors costs $O(rLD)$. The subspace descriptor
uses the same computation after replacing $\tilde K$ by its query-subspace projection $\tilde K U_Q$.

\clearpage
\section{Related Work}
\label{sec:related_work}

We focus on the query-aware, cacheable block selectors of \Cref{sec:background}, organized by the
\emph{cumulant order} of the cached summary. The broader landscape (fixed-pattern, KV-eviction,
low-rank) is surveyed in \Cref{sec:introduction}.

\subsection{CCQ}
\label{subsec:rw-ccq}
The closest related method is CCQ~\cite{ccq}, which rests on the same observation we make independently: a
log-partition (cumulant generating) function carries a second-order term set by a covariance, and
keeping it sharpens a first-order estimate. CCQ applies this to \emph{query correction} at read time
in linear attention; we instead apply it to \emph{block selection} in sparse attention, estimating
each block's softmax mass from the within-block key covariance $\Sigb$ (\Cref{sec:cumulant}). The
mathematics is shared, though the premises and resulting methods differ.

\subsection{First-order selectors}
\label{subsec:rw-firstmoment}
The prevailing cacheable block selectors summarize a block by a single cached vector. NSA~\cite{nsa}
scores each block from a learned MLP pooling of its keys (\Cref{sec:background}). DeepSeek-V4's
CSA~\cite{deepseekv4} pools each short token window by a learned, data-dependent weighting and selects
entries with a ReLU lightning indexer. Although the pooling can be nonlinear in the keys, the cached
entry is still query-independent and scored only to first order in $q$; the next subsection makes the
discarded covariance explicit.

\subsection{Limits of first-order selectors}
\label{subsec:rw-affine}
Consider first multi-head attention, where each head selects its own blocks. Selection ranks blocks
by normalized mass $P_b=\mass/Z$ with $\mass=\sum_{r\in b}e^{q\T k_r}$ (\Cref{sec:background}); the
shared denominator $Z$ makes this equivalent to ranking by $\mass$, hence by $\ln\mass$. A
first-order selector caches a single vector $\phi_b$ and an offline scalar $a_b$ per block and scores
affinely,
\begin{equation}
  \score_b^{\mathrm{aff}}(q) = a_b + q\T\phi_b .
  \label{eq:rw-affine}
\end{equation}
Cacheability is the sole constraint, enforcing that $\phi_b$ and $a_b$ be computed offline and independent of $q$;
they may, however, be arbitrary functions of the cache, not just a pool of the block's own keys. Therefore, 
mean-pooling ($\phi_b=\kb$, $a_b=\ln L$), NSA's MLP pooling ($\phi_b=\mathrm{MLP}(K_b)$, where
$K_b$ is the block's key matrix), and CSA's gated pooling are all of the same first-order class, represented by the affine score~\eqref{eq:rw-affine}. Note that the overlapping
windows of DeepSeek's NSA and CSA do not change this, as the cached vector stays a single 
query-independent vector.

\begin{figure}[t]
  \centering
  \includegraphics[width=0.925\linewidth]{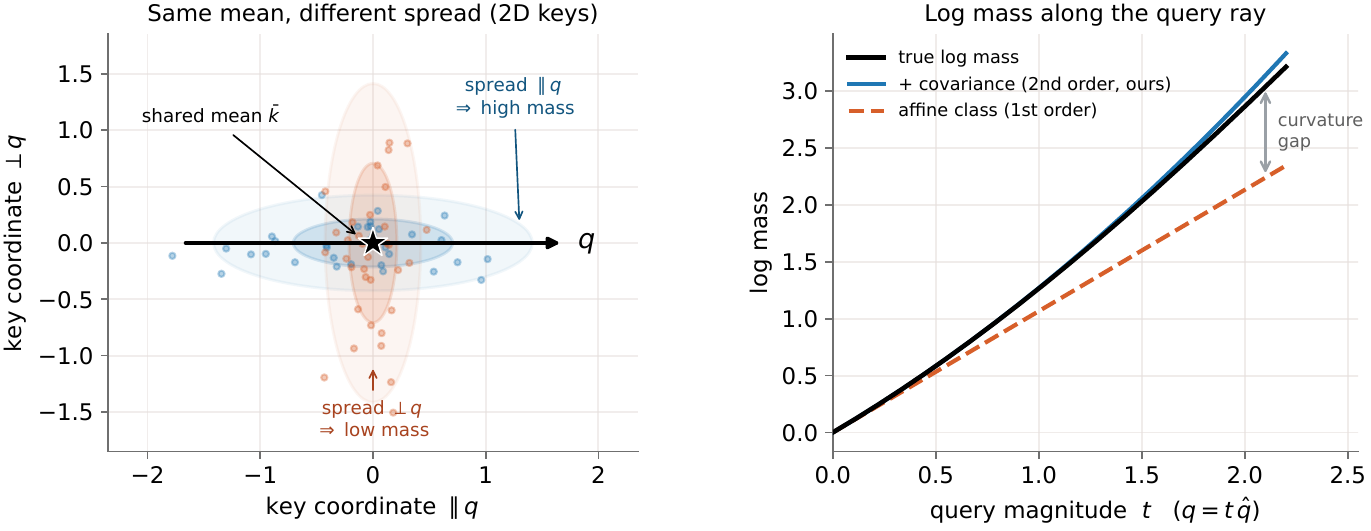}
  \caption{\textbf{First-order block scores miss within-block curvature.}
  \emph{Left (2D example):} two blocks share the same mean key $\kb$ but spread differently
  relative to the query direction $q$. Spread codirectional with $q$ (from the blue block) raises the block's softmax mass while
  spread orthogonal to $q$ (from the orange block) does not, a difference invisible to the shared mean $\kb$.
  \emph{Right:} for queries along a fixed direction $\hat q$, a first-order score such as $q^\top\kb$
  is linear in the query, whereas the true log mass is curved; the covariance term supplies the
  second-order curvature captured by COBS.}
  \label{fig:concept}
\end{figure}

The exact $\ln\mass$, however, is $\ln L$ plus the keys' cumulant generating function (\Cref{sec:cumulant}),
\begin{equation}
  \ln\mass = \ln L + q\T\kb + \tfrac12\,q\T\Sigb\,q + \cdots .
  \label{eq:rw-target}
\end{equation}
Choosing $a_b=\ln L$ and $\phi_b=\kb$ matches its constant and first-order terms, but no choice
reaches the quadratic $\tfrac12\,q\T\Sigb\,q$, which is not affine in $q$ (\Cref{fig:concept}).

Real GQA selectors share one block set across the $G$ query heads of a group and aggregate the
per-head masses (\Cref{eq:prop-gqa}), so the group score is a sum of per-head normalized masses
(softmax probabilities) and is no longer affine in $q$. One might hope this cross-head nonlinearity
recovers some curvature. We give an informal argument that it cannot lift the first-order family's
\emph{retrieval quality} past its ceiling. Sharing a single block set across heads is a
\emph{constraint} relative to letting each head select independently, and a more permissive selector
should only match or exceed the constrained one, giving the reasonable upper bound:
\begin{equation}
  \max_{\text{shared block sets}}\ \text{quality}
    \ \lesssim\ \max_{\text{independent block sets}}\ \text{quality} .
  \label{eq:rw-relax}
\end{equation}
Independent per-head selection reduces to the MHA case above, whose affine score is curvature-blind.
To the extent~\eqref{eq:rw-relax} holds, the retrieval quality of the whole first-order family,
including NSA's and CSA's selectors, is bounded by that curvature-blind ceiling. As such, this is an informal
bound on quality rather than a claim that the aggregated GQA score is itself affine.

This is corroborated by experiment, as the original NSA MLP scorer is the weakest selector we test
(\Cref{sec:experiments}): a more expressive pooling does not help, as its summary still enters the
score only to first order. CSA's ReLU indexer adds a query nonlinearity, yet it scores the same
query-independent entries to first order, so the same curvature-blindness applies.

\subsection{Beyond first-order selectors}
\label{subsec:rw-spread}
Escaping this first-order ceiling means caching more than a first-order summary, for instance a statistic of the
keys' spread. One example we analyze is Quest~\cite{quest}, which stores per block the element-wise
minimum and maximum of its keys,
\begin{equation}
  k_b^{\min} = \min_{r\in b} k_r, \qquad k_b^{\max} = \max_{r\in b} k_r \qquad(\text{element-wise}),
  \label{eq:rw-quest-box}
\end{equation}
an axis-aligned bounding box of the block's keys. It scores a block by the largest inner product any
point in that box could attain with the query,
\begin{equation}
  \hat s_b = \sum_i \max\!\big(q_i\,k_{b,i}^{\min},\ q_i\,k_{b,i}^{\max}\big)
           = \sum_i \big(\max(q_i,0)\,k_{b,i}^{\max} + \min(q_i,0)\,k_{b,i}^{\min}\big),
  \label{eq:rw-quest-score}
\end{equation}
keeping the top-scoring blocks. The min/max range is a coarse, axis-aligned proxy for how the keys
are spread, making $\hat s_b$ piecewise-linear rather than affine in $q$ and letting it carry \emph{some} of
the discarded curvature. Our method is the principled form: in place of a per-axis range we cache the
within-block covariance $\Sigb$ and score with the exact second-order term $\tfrac12\,q\T\Sigb\,q$ of
\eqref{eq:rw-target}.

\subsection{Orthogonal directions}
\label{subsec:rw-orthogonal}
Many efficient-attention directions are orthogonal to block selection; within the DeepSeek lineage
specifically, two sit just outside our scope. DeepSeek-V3.2's DSA~\cite{deepseekv32} selects
individual tokens with a lightweight ReLU ``lightning indexer'' rather than summarizing blocks, so the
cumulant-order analysis does not apply. DeepSeek-V4's HCA~\cite{deepseekv4} compresses long spans
(roughly $128$ tokens per entry) and attends densely, with no selection. Both reduce a different cost
axis and could compose with a better selector.

\section{Experiments}
\label{sec:experiments}

\subsection{Setup}
\label{subsec:exp-setup}
We use a $\approx$1.2B-parameter decoder-only transformer with $16$ layers, model dimension $2048$,
a SwiGLU FFN with width $8192$, $16$ query heads, $4$ KV heads under GQA, head dimension $D=128$, RoPE~\cite{rope} with base
$\theta=10^6$, and a $\approx$50k-token vocabulary. Pretraining uses LongCrawl64~\cite{longcrawl64}
for $\approx$20B tokens roughly following Chinchilla scaling~\cite{chinchilla} at a 4k sequence length. We then extend the context to 32k with
YaRN~\cite{yarn} and perform supervised fine-tuning (SFT) on generated RULER-style long-context data,
separate from the evaluation data (\Cref{app:sft}).

We evaluate long-context retrieval on RULER~\cite{ruler} at 32k, using 11 tasks: single-key, multi-key, multi-query, and multi-value needle retrieval; common-word and frequent-word
extraction (CWE, FWE); and variable tracking (VT). We also report standard zero-shot
common-sense benchmarks at this scale (\Cref{subsec:shortctx}) along with position-wise language-modeling loss (\Cref{subsec:posnll}). NSA branches are controlled to use $L=32$-token blocks, top-$k=16$ selected blocks, and a $256$-token sliding window.

We compare selectors under this shared protocol as separate controlled runs. \emph{NSA MLP} is our
NSA baseline: a controlled replication of DeepSeek's NSA~\cite{nsa}, which summarizes each block with a
generous $LD \to 4LD \to D$ MLP (ReLU activations) applied separately to keys and values, i.e.\ two
such MLPs. \emph{NSA mean-pool}
is a diagnostic baseline that replaces the learned MLP summary with the block
mean key, motivated by the cumulant expansion. \emph{NSA Quest} adapts Quest's~\cite{quest} per-page
key min/max scoring to the NSA selection branch. \emph{COBS} adds the within-block covariance summary instead, and our deliverable configuration is COBS ($s{\approx}85$, rank-$r{=}4$, FP4), with
full-space variants reported as ablations. NSA mean-pool, NSA Quest, and COBS share the same mean-pool
compression branch and differ only in the \emph{selection} scoring summary (block mean, Quest's
min/max, or COBS's covariance), so differences among them are selection-branch effects.

NSA MLP is the only parameter-count exception: at $\approx$1.7B we scale its pretraining token budget
proportionally to $\approx$28B tokens, versus $\approx$20B for all other variants at $\approx$1.2B. Subspace and FP4 results reuse the
corresponding full-space COBS checkpoint as inference-time changes to the selector descriptor and
scoring path, with no weight updates.

\subsection{Headline results (32k RULER)}
\label{subsec:headline}
\begin{figure}[H]
  \centering
  \definecolor{cobsbarfill}{RGB}{88,164,72}     
  \definecolor{cobsbarline}{RGB}{42,101,37}
  \definecolor{otherbarfill}{RGB}{170,178,189}  
  \definecolor{otherbarline}{RGB}{110,118,130}
  \definecolor{rulergrid}{RGB}{224,230,236}
  \begin{tikzpicture}
    \begin{axis}[
      ybar,
      width=\linewidth, height=6cm,
      ymin=0, ymax=1.0,
      ylabel={32k RULER},
      symbolic x coords={MLP/RoPE, mean/RoPE, mean/NoPE, NSA Quest, cobssub, cum4, cum6, OSA, dense},
      xtick={MLP/RoPE, mean/RoPE, mean/NoPE, NSA Quest, cobssub, cum4, cum6, OSA, dense},
      xticklabels={{MLP\\RoPE}, {mean\\RoPE}, {mean\\NoPE}, {NSA\\Quest}, {COBS\\subspace FP4}, {COBS full\\$r{=}4$}, {COBS full\\$r{=}6$}, {OSA\\mass oracle}, dense},
      x tick label style={font=\footnotesize, align=center, anchor=north},
      bar width=13pt,
      enlarge x limits=0.1,
      ymajorgrids,
      grid style={rulergrid},
      axis line style={black!55},
      tick style={black!55},
      nodes near coords style={font=\footnotesize, text=black},
    ]
      \addplot+[
        fill=otherbarfill, draw=otherbarline, line width=0.45pt, bar shift=0pt,
        nodes near coords, point meta=explicit symbolic
      ] coordinates {
        (MLP/RoPE,0.2999) [0.2999]
        (mean/RoPE,0.4186) [0.4186]
        (mean/NoPE,0.5554) [0.5554]
        (NSA Quest,0.5765) [0.5765]
        (cobssub,0) []
        (cum4,0) []
        (cum6,0) []
        (OSA,0.9010) [0.9010]
        (dense,0.9040) [0.9040]
      };
      \addplot+[
        fill=cobsbarfill, draw=cobsbarline, line width=0.45pt, bar shift=0pt,
        nodes near coords, point meta=explicit symbolic
      ] coordinates {
        (MLP/RoPE,0) []
        (mean/RoPE,0) []
        (mean/NoPE,0) []
        (NSA Quest,0) []
        (cobssub,0.8195) [0.8195]
        (cum4,0.8238) [0.8238]
        (cum6,0.8493) [0.8493]
        (OSA,0) []
        (dense,0) []
      };
    \end{axis}
  \end{tikzpicture}
  \caption{32k RULER gap ladder (COBS variants in green). The main gain comes from the
  covariance summary; the two mean-pool bars show the additive NoPE improvement over the RoPE baseline.}
  \label{fig:gapladder}
\end{figure}
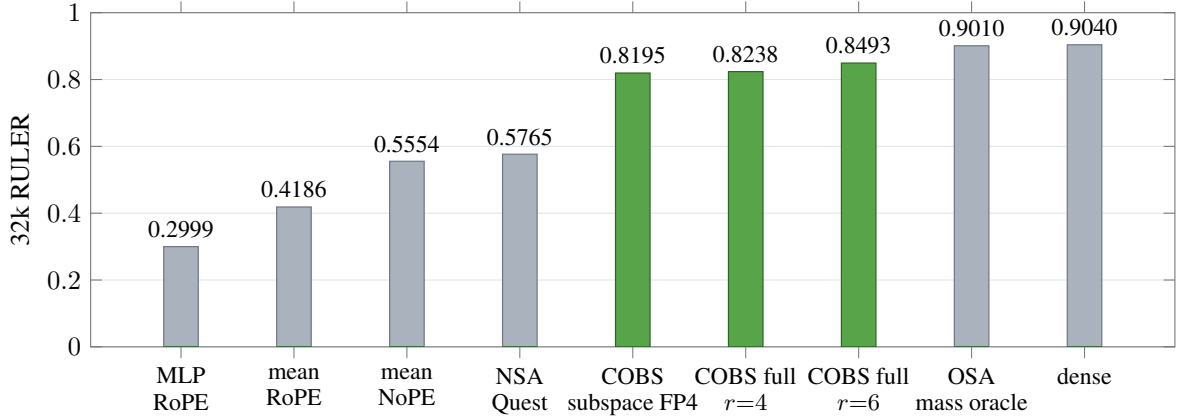

\Cref{fig:gapladder} and \Cref{tab:ruler-breakdown} show the headline result: extending first-order
selection with a covariance summary gives the main selector gain. COBS (subspace $s{\approx}85$,
$r{=}4$, FP4) reaches $0.8195$ on 32k RULER, closing about $86\%$ of the headroom from
the NSA MLP baseline ($0.2999$) to full (dense) attention ($0.9040$); the full-space low-rank descriptor
reaches $0.8238$ at $r=4$ and $0.8493$ at $r=6$ (\Cref{tab:ruler-breakdown}).

Removing RoPE from the selection summary is an additive improvement that raises the mean-pool baseline from $0.4186$ (RoPE) to $0.5554$ (NoPE), so we evaluate the NoPE-based selectors (NSA Quest,
COBS, and OSA) under this scheme. NSA Quest ($0.5765$) improves only modestly over mean-pool, while
COBS's covariance summary accounts for the large remaining gain.

\begin{table}[H]
  \caption{11-task 32k RULER breakdown. Gap closed is measured from NSA MLP (RoPE) ($0.2999$) to full (dense) attention ($0.9040$).
  NSA MLP (RoPE) is the native baseline, and we note that the NoPE scheme does not improve the NSA MLP baseline. Subtask scores are rounded.}
  \label{tab:ruler-breakdown}
  \centering
  \definecolor{cumulantrowlight}{RGB}{231,244,224}%
  \definecolor{cumulantrowdeep}{RGB}{205,232,191}%
  \resizebox{\textwidth}{!}{%
  \setlength{\tabcolsep}{3pt}%
  \renewcommand{\arraystretch}{1.2}%
  \begin{tabular}{lccccccccccccc}
    \toprule
    Method & Mean & Gap closed & S1 & S2 & S3 & MK1 & MK2 & MK3 & MQ & MV & CWE & FWE & VT \\
    \midrule
    Dense (full attention)             & 0.9040 & 100.0\% & 1.00 & 1.00 & 1.00 & 0.98 & 0.95 & 0.92 & 0.92 & 0.93 & 0.44 & 0.90 & 0.92 \\
    OSA (mass oracle)                  & 0.9010 &  99.5\% & 1.00 & 1.00 & 1.00 & 0.99 & 0.94 & 0.91 & 0.96 & 0.96 & 0.22 & 0.95 & 0.98 \\
    \midrule
    NSA MLP (RoPE)                     & 0.2999 &   0.0\% & 1.00 & 0.11 & 0.05 & 0.09 & 0.00 & 0.00 & 0.09 & 0.10 & 0.10 & 0.93 & 0.83 \\
    NSA MLP (NoPE)                     & 0.2822 &  $-2.9$\% & 1.00 & 0.07 & 0.25 & 0.09 & 0.00 & 0.00 & 0.08 & 0.08 & 0.03 & 0.95 & 0.54 \\
    NSA mean-pool (RoPE)               & 0.4186 &  19.6\% & 1.00 & 0.35 & 0.62 & 0.22 & 0.14 & 0.00 & 0.34 & 0.30 & 0.03 & 0.90 & 0.72 \\
    NSA mean-pool (NoPE)               & 0.5554 &  42.3\% & 1.00 & 0.78 & 0.82 & 0.53 & 0.01 & 0.00 & 0.47 & 0.47 & 0.15 & 0.91 & 0.97 \\
    NSA Quest                          & 0.5765 &  45.8\% & 1.00 & 0.86 & 0.87 & 0.39 & 0.03 & 0.00 & 0.55 & 0.55 & 0.19 & 0.93 & 0.97 \\
    \midrule
    \rowcolor{cumulantrowlight} COBS full-space $r{=}4$ & 0.8238 &  86.7\% & 1.00 & 1.00 & 1.00 & 0.98 & 0.78 & 0.34 & 0.89 & 0.92 & 0.24 & 0.94 & 0.97 \\
    \rowcolor{cumulantrowlight} COBS full-space $r{=}6$ & 0.8493 &  90.9\% & 1.00 & 1.00 & 1.00 & 1.00 & 0.91 & 0.48 & 0.94 & 0.94 & 0.15 & 0.94 & 0.99 \\
    \rowcolor{cumulantrowdeep} COBS (subspace $s{\approx}85$, $r{=}4$, FP4) & 0.8195 &  86.0\% & 1.00 & 1.00 & 1.00 & 0.97 & 0.79 & 0.31 & 0.89 & 0.91 & 0.23 & 0.94 & 0.97 \\
    \bottomrule
  \end{tabular}}
\end{table}

\clearpage
\subsection{Short-context common-sense reasoning (parity)}
\label{subsec:shortctx}
\Cref{tab:shortctx} checks whether the sparse variants preserve short-context reasoning on
OpenBookQA~\cite{obqa}, PIQA~\cite{piqa}, HellaSwag~\cite{hellaswag}, ARC~\cite{arc},
TriviaQA~\cite{triviaqa}, and WinoGrande~\cite{winogrande}. These
benchmarks contain only tens to hundreds of tokens, so the local window plus top-$k$ selected
blocks already covers almost all of the input. In this regime, selection has little room to help:
the desired outcome is parity with dense attention, which all variants achieve within a $0.6$-point
average spread.

\begin{table}[H]
  \definecolor{cumulantrowlight}{RGB}{231,244,224}%
  \definecolor{cumulantrowdeep}{RGB}{205,232,191}%
  \caption{Short-context common-sense reasoning: zero-shot accuracy (\%) over seven tasks. At these
  sequence lengths the selection budget largely covers the input, and all sparse variants preserve
  dense-like performance. NSA MLP here refers to the baseline \emph{with} RoPE, while OSA, mean-pool, NSA Quest, and COBS use the NoPE scheme.}
  \label{tab:shortctx}
  \centering
  \resizebox{0.8\linewidth}{!}{%
  \setlength{\tabcolsep}{5pt}%
  \begin{tabular}{lcccccccc}
    \toprule
    Model & OBQA & PIQA & Hella. & ARC-c & TQA & ARC-e & Wino. & Avg. \\
    \midrule
    Dense                  & 18.8 & 63.6 & 31.1 & 23.1 & 42.5 & 37.3 & 50.7 & 38.2 \\
    OSA (mass oracle)      & 17.0 & 63.7 & 31.5 & 20.7 & 44.3 & 38.1 & 50.7 & 38.0 \\
    \midrule
    NSA mean-pool          & 17.2 & 63.0 & 31.3 & 21.8 & 43.1 & 37.9 & 51.9 & 38.0 \\
    NSA MLP                & 17.8 & 62.7 & 31.9 & 22.3 & 44.7 & 37.8 & 50.6 & 38.3 \\
    NSA Quest              & 16.0 & 64.7 & 31.4 & 21.5 & 44.6 & 38.4 & 52.2 & 38.4 \\
    \rowcolor{cumulantrowdeep} COBS & 17.8 & 64.5 & 31.3 & 23.5 & 44.0 & 38.3 & 50.9 & 38.6 \\
    \bottomrule
  \end{tabular}}
\end{table}

\subsection{Position-wise language-modeling loss}
\label{subsec:posnll}
\begin{figure}[H]
  \centering
  \includegraphics[width=1\linewidth]{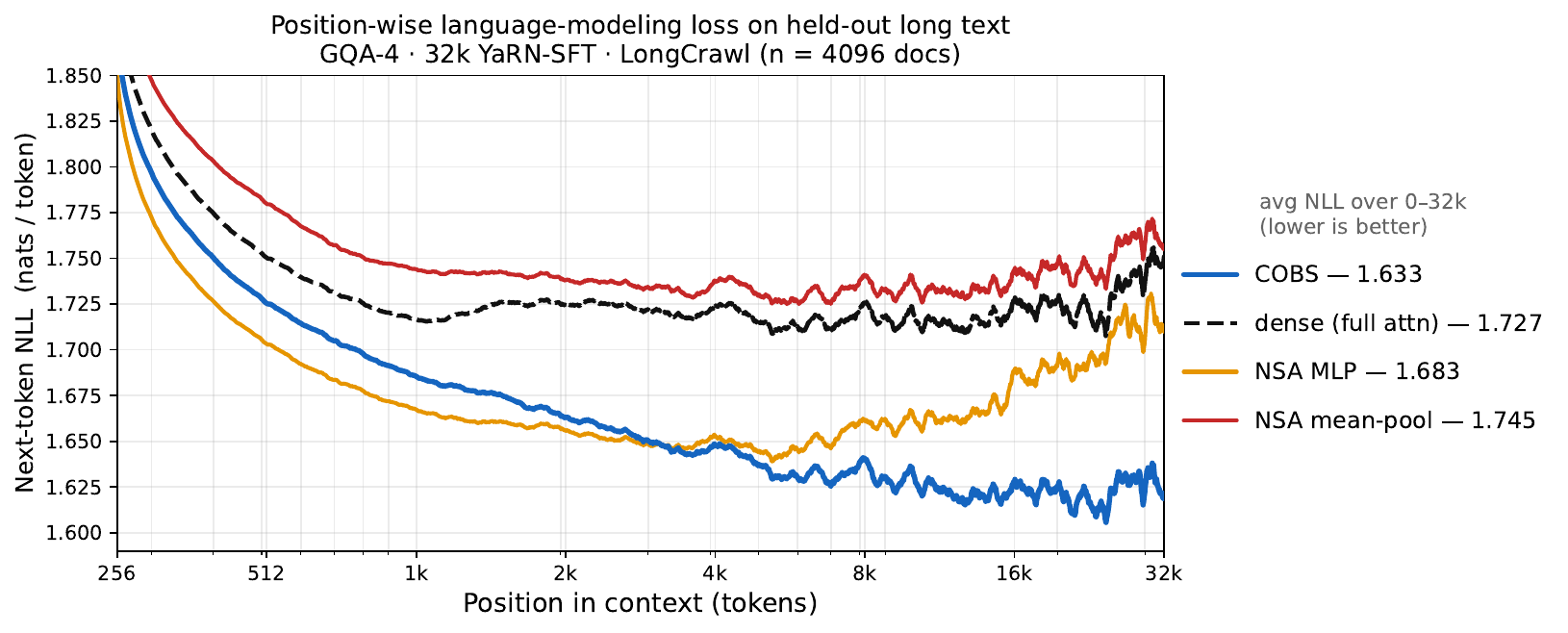}
  \caption{Position-wise next-token NLL on held-out natural LongCrawl64 at 32k (GQA-4, 32k YaRN SFT;
  $\approx$1.2B except NSA MLP at $\approx$1.7B).}
  \label{fig:posnll}
\end{figure}

RULER isolates retrieval, so we also evaluate ordinary next-token prediction over long natural
contexts. \Cref{fig:posnll} reports position-wise NLL for the SFT-trained GQA-4 variants.

COBS has the lowest average NLL ($1.633$), below dense ($1.727$), NSA MLP ($1.683$), and NSA
mean-pool ($1.745$). NSA MLP's lower short-context loss here is likely a
byproduct of our charitable baseline, because it carries many more parameters ($\approx$1.7B vs $\approx$1.2B) with a
much longer, proportionally scaled pretraining budget. Furthermore, the average
alone is weak evidence, since strong short-context prediction can mask a weak long-context tail. The more telling signal is the behavior at long positions, where
COBS's NLL slope stays flat and shows no upturn, unlike the weaker selectors. This indicates COBS
is conditioning on distant tokens rather than leaning on the local window, so the retrieval gains in
\Cref{subsec:headline} do not come at the cost of long-context language modeling.

\subsection{KV cache read traffic study}
\label{subsec:isoband}
We report the KV cache read traffic per decode step and layer
(\Cref{tab:bandwidth}). The accounting uses a 32k context with $H=4$ KV heads of dimension
$D=128$ and the same $L=32$, top-$k=16$, and $256$-token window from setup. Methods differ in
the per-block summary read to score blocks. Dense reads the whole KV cache, while OSA re-reads all
raw keys to compute the true attention masses and normalized GQA score before using the same
NSA read pattern. The resulting accuracy--traffic tradeoff is summarized in \Cref{fig:pareto}.

\begin{table}[H]
  \caption{Per-decode-step, per-layer KV cache read traffic at 32k, by branch (KiB;
  $1\,\mathrm{KiB}=1024$ bytes). Summary keys: each method's per-block key summary read. Summary values: the compression branch's per-block value summary read. Window: the
  $256$-token local window. Fine-grained: full K,V for the top-$k=16$ selected blocks (the
  full-cache read for Dense). Last two columns report dense read traffic divided by each method's
  traffic ($\times$less), and each method's traffic divided by NSA MLP's traffic ($3584$ KiB; $\times$more).
  Note that NSA Quest is a modification of NSA mean-pool's selection branch only.}
  \label{tab:bandwidth}
  \definecolor{cumulantrowlight}{RGB}{231,244,224}%
  \definecolor{cumulantrowdeep}{RGB}{205,232,191}%
  \centering
  \resizebox{\linewidth}{!}{%
  \renewcommand{\arraystretch}{1.2}%
  \begin{tabular}{lccccccc}
    \toprule
    Method & Summary keys & Summary values & Window & Fine grained & Per layer & vs.\ dense & vs.\ NSA MLP \\
           & (KiB)        & (KiB)          & (KiB)  & (KiB)        & (KiB)     & ($\times$less) & ($\times$more) \\
    \midrule
    Dense (full attention)         & --         & --     & --    & $65{,}536$ & $65{,}536$ & -- & $18.29\times$ \\
    OSA (mass oracle)              & $33{,}792$ & $1024$ & $512$ & $1024$     & $36{,}352$ & $1.80\times$ & $10.14\times$ \\
    \midrule
    NSA MLP                        & $1024$     & $1024$ & $512$ & $1024$     & $3584$     & $18.29\times$ & -- \\
    NSA mean-pool                  & $1024$     & $1024$ & $512$ & $1024$     & $3584$     & $18.29\times$ & $1.00\times$ \\
    NSA Quest                      & $3072$     & $1024$ & $512$ & $1024$     & $5632$     & $11.64\times$ & $1.57\times$ \\
    \midrule
    COBS full-space $r{=}4$ (bf16)   & $5120$     & $1024$ & $512$ & $1024$     & $7680$     & $8.53\times$  & $2.14\times$ \\
    COBS full-space $r{=}6$ (bf16)   & $7168$     & $1024$ & $512$ & $1024$     & $9728$     & $6.74\times$  & $2.71\times$ \\
    \rowcolor{cumulantrowlight} COBS full-space $r{=}4$ (FP4)    & $2112$     & $1024$ & $512$ & $1024$     & $4672$     & $14.03\times$ & $1.30\times$ \\
    \rowcolor{cumulantrowlight} COBS full-space $r{=}6$ (FP4)    & $2656$     & $1024$ & $512$ & $1024$     & $5216$     & $12.56\times$ & $1.46\times$ \\
    \rowcolor{cumulantrowdeep} COBS (subspace $s{\approx}85$, $r{=}4$, FP4) & $1767$ & $1024$ & $512$ & $1024$ & $4327$ & $15.15\times$ & $1.21\times$ \\
    \bottomrule
  \end{tabular}}
\end{table}

\subsection{Ablations}
\label{subsec:ablations}

\subsubsection{Rank sweep and high-rank regression}
\begin{figure}[H]
  \centering
  \vspace{-0.35em}
  \includegraphics[width=0.70\linewidth]{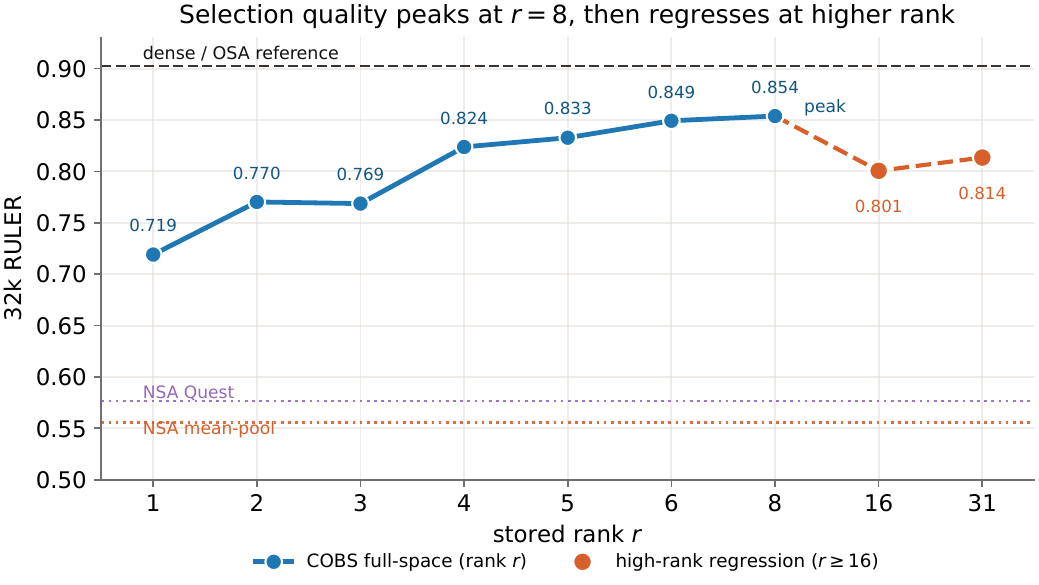}
  \vspace{-0.65em}
  \caption{COBS full-space rank sweep on 32k RULER. Selection peaks at $r=8$ ($0.8539$), then regresses at
  $r=16$ ($0.8006$) and stays below the peak at the maximum rank $r=31$ ($0.8135$).}
  \label{fig:ranksweep}
  \vspace{-0.45em}
\end{figure}

Selection rises with the stored rank through $r=8$, which peaks at $0.8539$ on 32k RULER, but then
\emph{regresses} to $0.8006$ at $r=16$ (\Cref{fig:ranksweep}). The collapse is concentrated in the multi-key needle subtasks: MK3 falls from $0.470$ at $r=8$ to $0.054$ at $r=16$ and MK2 from $0.934$ to $0.800$,
while the single-needle subtasks stay saturated ($1.000$).

We attribute this to the \emph{unsigned} variance term in the covariance summary. To second order, the
score combines signed alignment, $q^\top\bar k$, with the nonnegative curvature term
$\tfrac12\,q^\top\Sigma q$. Additional eigenvectors can therefore boost blocks whose
keys have large variance along the query direction even when the aligned signal is weak or oppositely
signed, for instance, blocks with many distractors. By $r=16$, this false-positive mass overwhelms the genuine signal on multi-key tasks.
This does not reverse at higher rank: with $L=32$ keys per block the covariance has rank at most $L-1=31$, yet the full $r=31$
descriptor still scores only $0.8135$, below the $r=8$ peak, with MK3 still collapsed ($0.062$); retaining eigenvectors beyond the few
dominant directions does not recover the signal. The
low-rank descriptors ($r=4$--$8$) avoid much of this effect by retaining only the few dominant
directions, so we keep $r\le 8$ as the operating range.

\subsubsection{Subspace: adaptive vs.\ global}
\begin{figure}[H]
  \centering
  \definecolor{cobsbarfill}{RGB}{88,164,72}%
  \definecolor{cobsbarlight}{RGB}{176,212,163}%
  \definecolor{cobsbarline}{RGB}{42,101,37}%
  \definecolor{rulergrid}{RGB}{224,230,236}%
  \begin{tikzpicture}
    \begin{axis}[
      ybar stacked,
      width=\linewidth, height=5cm,
      ymin=0, ymax=120,
      xlabel={Layer},
      ylabel={Subspace dim.\ $s$},
      xtick={1,2,3,4,5,6,7,8,9,10,11,12,13,14,15,16},
      x tick label style={font=\footnotesize},
      bar width=11pt,
      enlarge x limits=0.04,
      ymajorgrids,
      grid style={rulergrid},
      axis line style={black!55},
      tick style={black!55},
    ]
      \addplot+[fill=cobsbarfill, draw=cobsbarline, line width=0.4pt] coordinates {
        (1,34) (2,45) (3,59) (4,56) (5,50) (6,71) (7,75) (8,77)
        (9,86) (10,78) (11,79) (12,80) (13,79) (14,77) (15,72) (16,70)
      };
      \addplot+[fill=cobsbarlight, draw=cobsbarline, line width=0.4pt] coordinates {
        (1,8) (2,11) (3,14) (4,15) (5,12) (6,18) (7,19) (8,19)
        (9,22) (10,20) (11,19) (12,19) (13,20) (14,19) (15,18) (16,17)
      };
      \draw[dashed, black!75, line width=0.8pt]
        (axis cs:0.4,68) -- (axis cs:16.6,68);
      \draw[dashed, black!65, line width=0.8pt]
        (axis cs:0.4,84.875) -- (axis cs:16.6,84.875);
      \node[font=\footnotesize, black!75, anchor=south west] at (axis cs:0.6,68) {avg $=68$};
      \node[font=\footnotesize, black!65, anchor=south west] at (axis cs:0.6,84.875) {avg $=84.875$};
    \end{axis}
  \end{tikzpicture}
  \caption{Adaptive per-layer query-subspace dimension $s$ for the deliverable COBS configuration
  ($r{=}4$, FP4). Each bar's lower (dark) portion is the rank capturing $90\%$ of the query spectral
  energy, averaged over heads (across-layer average $s\approx68$); the full bar scales this by
  $1.25\times$ to give the deliverable configuration (average $s=84.875$, i.e.\ $s\approx85$). Dashed
  lines mark the two across-layer averages. Early layers need far fewer query dimensions; the
  $s\approx85$ average is the value underlying \Cref{tab:subspace,tab:bandwidth}.}
  \label{fig:subspace-perlayer}
\end{figure}
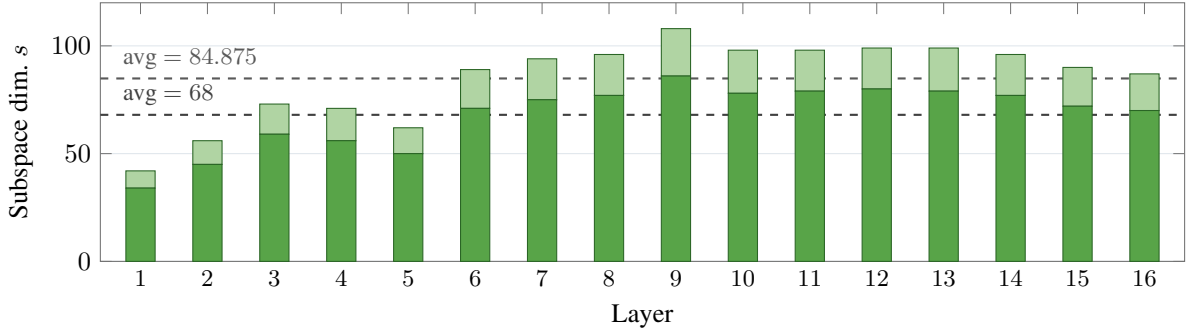

Adaptive per-layer subspaces outperform a single global budget at comparable dimension. At an
average of $s\approx85$, the score is within ${\approx}0.005$ of the full low-rank form
($s=128$; \Cref{tab:subspace}). Concretely, the $s\approx85$ configuration sets each layer's $s$ to
$1.25\times$ the rank capturing $90\%$ of the query spectral energy, averaged over heads; the resulting allocation
(\Cref{fig:subspace-perlayer}) concentrates dimensions in the later layers, while the early layers
need far fewer.

\begin{table}[H]
  \definecolor{cumulantrowlight}{RGB}{231,244,224}%
  \definecolor{cumulantrowdeep}{RGB}{205,232,191}%
  \caption{Global (blanket) versus adaptive per-layer subspace dimension $s$ on 32k RULER, applied at
  inference time to the full-space low-rank $r=4$ checkpoint. Global $s=128$ is the full low-rank form
  (no subspace reduction); the adaptive rows set $s$ per layer and report the across-layer average.}
  \label{tab:subspace}
  \centering
  \begin{tabular}{llc}
    \toprule
    Allocation & $s$ (dim) & 32k RULER \\
    \midrule
    Global               & 128 (full)         & 0.8238 \\
    Global               & 96                 & 0.8188 \\
    Global               & 64                 & 0.7856 \\

    \midrule
    \rowcolor{cumulantrowdeep} Adaptive (per-layer) & ${\approx}85$ (avg) & 0.8195 \\
    Adaptive (per-layer) & ${\approx}68$ (avg) & 0.8054 \\
    \bottomrule
  \end{tabular}
\end{table}

\subsubsection{Descriptor quantization}
\label{subsec:quant-ablation}
\Cref{tab:quant} quantizes the stored scaled eigenvectors $\xi_i$ while keeping the block mean at
higher precision. FP4 (E2M1) is the COBS format.

\begin{table}[H]
  \definecolor{cumulantrowlight}{RGB}{231,244,224}%
  \definecolor{cumulantrowdeep}{RGB}{205,232,191}%
  \caption{Descriptor quantization on 32k RULER; ``bytes/desc'' counts covariance-factor bytes only,
  excluding the bf16 block mean, and ``reduction'' is versus the bf16 descriptor. FP4 is essentially
  lossless.}
  \label{tab:quant}
  \centering
  \small
  \begin{tabular}{llcccc}
    \toprule
    Descriptor & Quant & 32k RULER & $\Delta$ vs bf16 & bytes/desc & reduction \\
    \midrule
    Full-space $r{=}4$ (128-dim) & bf16                    & 0.8238 & ---      & 1024 & $1\times$ \\
    \rowcolor{cumulantrowdeep} & fp4 (E2M1)              & 0.8251 & $+0.0013$ & 272  & $3.8\times$ \\
    \midrule
    Subspace $r{=}4, s{=}64$ (64-dim) & bf16                    & 0.7856 & ---       & 512  & $1\times$ \\
    \rowcolor{cumulantrowdeep}                           & fp4 (E2M1)              & 0.7844 & $-0.0012$ & 144  & $3.6\times$ \\
    \midrule
    Full-space $r{=}6$ (128-dim) & bf16                    & 0.8493 & ---      & 1536 & $1\times$ \\
    \rowcolor{cumulantrowdeep} & fp4 (E2M1)              & 0.8467 & $-0.0026$ & 408  & $3.8\times$ \\
    \bottomrule
  \end{tabular}
\end{table}

\subsection{Negative results}
\label{subsec:negative}
We include two negative results because each motivates a design choice.

\paragraph{Query-centered expansion.} Expanding the CGF around a calibrated query origin $q_0$
instead of $q=0$ underperforms, for two reasons. First, under the same cache budget, the tilted
moments must be shared by the $G$ query heads in a KV group. Second, the approximation is local
around $q_0$ and can degrade on outlier needle queries far from that origin. Empirically it lowers
32k RULER from $0.8238$ to $0.8100$ (\Cref{tab:querycentered}), concentrated on the multi-key and
multi-value needle tasks (MK3 $0.34\!\to\!0.26$, MV $0.92\!\to\!0.88$). We therefore expand at
$q=0$.

\begin{table}[H]
  \caption{Query-centered expansion on 32k RULER (COBS full-space $r{=}4$): expanding the CGF around a calibrated
  origin $q_0$ instead of $q=0$ regresses selection, concentrated on the multi-key and multi-value
  needle tasks.}
  \label{tab:querycentered}
  \centering
  \resizebox{0.80\textwidth}{!}{%
  \setlength{\tabcolsep}{3pt}%
  \begin{tabular}{lcccccccccccc}
    \toprule
    Method & Mean & S1 & S2 & S3 & MK1 & MK2 & MK3 & MQ & MV & CWE & FWE & VT \\
    \midrule
    COBS full-space $r{=}4$ ($q{=}0$)        & 0.8238 & 1.00 & 1.00 & 1.00 & 0.98 & 0.78 & 0.34 & 0.89 & 0.92 & 0.24 & 0.94 & 0.97 \\
    \quad query-centered ($q_0$) & 0.8100 & 1.00 & 1.00 & 1.00 & 0.97 & 0.75 & 0.26 & 0.88 & 0.88 & 0.25 & 0.94 & 0.97 \\
    \bottomrule
  \end{tabular}}
\end{table}

\paragraph{Cheap diagonal skew.} As the variance term is sign-blind (\Cref{fig:ranksweep}), a
natural fix is adding the signed third cumulant $\kappa_3$. Caching the full $O(D^3)$ third-order tensor is
impractical, but the diagonal in the stored eigenbasis $u_i$ costs one scalar $g_i$ per eigenvector
and adds a signed cubic correction to the block log-score,
\begin{equation}
  \tfrac16\!\sum_{ijk}(\kappa_3)_{ijk}\,q_iq_jq_k
  \;\approx\; \tfrac16\sum_{i=1}^{r}(u_i\T q)^3\,g_i,
  \qquad
  g_i=\frac1L\sum_{t\in b}\big(u_i\T(k_t-\kb)\big)^3 .
\end{equation}
Empirically the effect depends on the stored rank (\Cref{tab:skew}). At low rank the correction only
\emph{hurts}: the cheap skew approximation drops $r=4$ from $0.8238$ to $0.7754$, and $r=6$ from
$0.8493$ to $0.7696$, in both cases collapsing the multi-key needles (at $r=4$,
MK2 $0.78\!\to\!0.54$, MK3 $0.34\!\to\!0.01$). The diagonal projection discards mixed third-order
interactions and is dominated by heavy-tailed outlier keys, thereby confounding the retained dominant eigenvectors.

However, the signed term helps in the regressed high-rank regime of \Cref{fig:ranksweep}.
At $r=16$, adding sixteen skew scalars raises 32k RULER from $0.8006$ (no skew) to $0.8252$ and
partially undoes the multi-key collapse (MK3 $0.05\!\to\!0.34$). This supports our diagnosis: the extra eigenvectors accumulate spurious \emph{unsigned} variance mass on
blocks with many distractors, and the signed cubic term ameliorates this by canceling part of that false-positive mass. Even so, the repaired high-rank point ($0.8252$) still trails the clean low-rank
covariance operating range (peak $0.8539$ at $r=8$, $0.8493$ at $r=6$) while paying extra KV cache read traffic. We therefore stop at the covariance.

\begin{table}[H]
  \caption{Cheap diagonal skew on 32k RULER: adding $r$ skew scalars $g_i$ along the stored
  eigendirections. The signed cubic term regresses at low rank ($r=4,6$), collapsing the multi-key
  needles, but at $r=16$ it partially repairs the high-rank regression (\Cref{fig:ranksweep}), lifting
  the mean and recovering MK3.}
  \label{tab:skew}
  \centering
  \resizebox{0.80\textwidth}{!}{%
  \setlength{\tabcolsep}{3pt}%
  \begin{tabular}{lcccccccccccc}
    \toprule
    Method & Mean & S1 & S2 & S3 & MK1 & MK2 & MK3 & MQ & MV & CWE & FWE & VT \\
    \midrule
    COBS full-space $r{=}4$          & 0.8238 & 1.00 & 1.00 & 1.00 & 0.98 & 0.78 & 0.34 & 0.89 & 0.92 & 0.24 & 0.94 & 0.97 \\
    \quad + diagonal skew ($4$)      & 0.7754 & 1.00 & 1.00 & 1.00 & 0.97 & 0.54 & 0.01 & 0.93 & 0.94 & 0.20 & 0.97 & 0.96 \\
    \midrule
    COBS full-space $r{=}6$          & 0.8493 & 1.00 & 1.00 & 1.00 & 1.00 & 0.91 & 0.48 & 0.94 & 0.94 & 0.15 & 0.94 & 0.99 \\
    \quad + diagonal skew ($6$)      & 0.7696 & 1.00 & 1.00 & 1.00 & 0.97 & 0.60 & 0.02 & 0.89 & 0.93 & 0.16 & 0.94 & 0.97 \\
    \midrule
    COBS full-space $r{=}16$         & 0.8006 & 1.00 & 1.00 & 1.00 & 0.96 & 0.80 & 0.05 & 0.92 & 0.94 & 0.25 & 0.91 & 0.98 \\
    \quad + diagonal skew ($16$)     & 0.8252 & 1.00 & 1.00 & 1.00 & 0.99 & 0.80 & 0.34 & 0.93 & 0.94 & 0.17 & 0.94 & 0.97 \\
    \bottomrule
  \end{tabular}}
\end{table}

\section{Conclusion}
\label{sec:conclusion}

Block sparsity is arguably the most hardware friendly form of sparse attention, yet it remains conspicuously absent from leading open-weight LLMs.
Through the lens of NSA, we traced the present challenges of block sparse methods to the selection branch, and showed that selection
reduces to ranking blocks by their attention mass. An oracle that ranks by the exact mass essentially
matches dense attention, therefore reducing the remaining problem to precisely estimating the mass from a cacheable, query-independent summary.

A cumulant expansion of the block mass illustrates the precise limitations of existing selectors: their cached
summaries provide only a score that is first-order in the query, discarding the within-block key
covariance that supplies the second-order curvature term $\tfrac12 q\T\Sigb q$. COBS keeps the
cacheability constraint while raising the cumulant order of the summary, storing a compressed
second-order statistic per block.

Empirically, raising the cumulant order with a compressed covariance yields a large gain: COBS (subspace
$s{\approx}85$, $r{=}4$, FP4) lifts our 11-task 32k RULER score from the mean-pool baseline's $0.5554$ to $0.8195$.
Measured against the full headroom, these additive changes close about $86\%$ of the gap between the NSA MLP
baseline ($0.2999$) and full (dense) attention ($0.9040$). The same model preserves short-context
common-sense performance and attains the lowest position-wise language-modeling NLL in our comparison
($1.633$, versus dense at $1.727$) while using only $1.21\times$ the NSA baseline's KV cache read traffic
and $15.15\times$ less than dense.

Cumulant order is therefore both a diagnostic lens for existing cacheable selectors and a practical design axis for better methods, and we see it as a step towards making block sparse attention a more prevalent strategy.

\section{Limitations}
\label{sec:limitations}

\paragraph{Scale.} This is a mechanism study at $\approx$1.2B backbone scale with a 4k pretraining
sequence length; the NSA MLP replication is $\approx$1.7B after adding its learned selector. The results
characterize \emph{why} second-order selection helps, rather than demonstrating a deployment-scale system.

\paragraph{Controlled NSA comparison.} Our setup differs from the original NSA: we use different
hyperparameters, a large MLP for the underspecified compression branch, and non-overlapping blocks. These
are held fixed across the selector variants, which are separate controlled runs. Our numbers should therefore be read as system-level
comparisons against a strong NSA baseline, not as a reproduction of NSA's reported results.

\paragraph{NoPE confound.} The sparse variants remove RoPE from the compression and selection
branches (the NoPE scheme, \Cref{subsec:nope}), which we found benefits long-context behavior, whereas
the dense baseline retains RoPE throughout. Long-context comparisons between the sparse variants and
dense therefore partly reflect this position-encoding difference and not sparsity alone. Moreover, NoPE
was only ablated on our long-context retrieval configuration, where the content of tokens may have had
higher importance than their relative position.

\paragraph{RULER-style SFT.} Our long-context signal comes from fine-tuning on generated RULER-style
data (\Cref{app:sft}), a nonstandard protocol whose rankings may not match a more
representative long-context setting. We therefore read our RULER numbers as \emph{relative} selection
quality rather than absolute accuracy. The comparison is still informative because the protocol is
identical and generous for every selector: the first-order NSA baseline falls well below dense
even under this favorable SFT, and OSA matches dense while COBS recovers most of the gap,
illustrating meaningful differences in selection quality. Relatedly, the upward long-context slope in
the position-wise NLL of some variants (\Cref{subsec:posnll}) may itself be an artifact of the
RULER-style SFT.

\paragraph{KV read accounting.} COBS stores more per block than the mean-pool baseline: without
quantization it costs more KV cache read traffic, and it still does if standard KV is also kept in
FP4. In all cases the footprint stays far below dense, and FP4 quantization of the stored eigenvectors
is empirically lossless on our evaluation (\Cref{subsec:quant}). These accounting numbers do not by
themselves imply end-to-end runtime gains, which depend on kernels, batching, hardware, and decoding regime.

\section*{Contributions and Acknowledgments}

Alexander Tian and Aditya Ghai developed the primary analysis and methods, including the selection-oracle derivation, the cumulant expansion, the cumulant-order analysis, and the compression approaches; they conducted the experiments and wrote the paper. Sanjit Neelam proposed and implemented SFT on RULER-like examples to assess long-context ability under a limited compute budget, and wrote the kernels used to train the NSA baselines. Sanjit Neelam and Zaal Vasania developed the NoPE scheme for the compression and selection branches, among other prior experiments that provided useful insights. Akshay Mishra developed OSA and empirically found that it closes most of the gap between block sparse attention and dense attention; he also set the research direction of approximating OSA without reading all keys, and provided senior guidance throughout the project.

We also thank Vaclav Cvicek and Daniel Heinlein for their much appreciated feedback on earlier drafts.

We additionally thank Sanjit Neelam, Zaal Vasania, Akshay Mishra, Vaclav Cvicek, Hayden Le, Daniel Heinlein, Neil Adit, and Reiner Pope for developing and maintaining the MatX
training and evaluation infrastructure used for these experiments. We use seqax~\cite{seqax2024github}, MatX's
research-focused LLM codebase built on JAX, to perform all experiments.

\bibliographystyle{unsrtnat}
\bibliography{references}

\clearpage
\appendix

\section{Derivation of the selection score}
\label{app:selection-derivation}

We derive the additive GQA selection score~\eqref{eq:prop-gqa} and its MHA special case from the
per-KV-head objective~\eqref{eq:oracle-multihead} under \Cref{ass:value-blind,ass:gamma,ass:linear}.

\paragraph{Reducing the objective under A1.}
Applying the triangle inequality over the dropped blocks in~\eqref{eq:oracle-exact} and using
\Cref{ass:value-blind} bounds each query head's reconstruction error:
\begin{equation}
  \big\lVert(\ostar-\ohat_S)^{(g,h)}\big\rVert
  \le\frac{1}{1-\tau^{(g,h)}}\sum_{b\in S^c}P_b^{(g,h)}\big\lVert(\vcb-\ostar)^{(g,h)}\big\rVert
  \le c^{(g,h)}\,\frac{\tau^{(g,h)}}{1-\tau^{(g,h)}}.
  \label{eq:oracle-a1-head}
\end{equation}
Summing over the group bounds the per-head objective~\eqref{eq:oracle-multihead},
\begin{equation}
  \mathcal{E}^{(h)}(S)\;\le\;\sum_{g=1}^{G} c^{(g,h)}\,\frac{\tau^{(g,h)}}{1-\tau^{(g,h)}}.
  \label{eq:oracle-residual-a1}
\end{equation}

\paragraph{The GQA score.}
For GQA ($G>1$), disregarding $c^{(g,h)}$ (\Cref{ass:gamma}) leaves the per-KV-head objective
\eqref{eq:oracle-residual-a1} as $\min_{S^{(h)}}\sum_g\tau^{(g)}/(1-\tau^{(g)})$ (fixing the KV head
$h$, we abbreviate $\tau^{(g,h)}$ as $\tau^{(g)}$). This is set-dependent: the $1/(1-\tau^{(g)})$
factor makes the gain from keeping a block depend on which blocks are already kept, so no fixed
per-block score is exactly optimal. Performing a Taylor expansion of the penalty centered at 0,
\[
  \frac{\tau^{(g)}}{1-\tau^{(g)}}=\tau^{(g)}+O\!\big((\tau^{(g)})^2\big),
\]
and replacing it by its linear leading term (\Cref{ass:linear}) makes the objective additive, hence
maximized by the top-$k$ blocks under the additive score~\eqref{eq:prop-gqa}.

\paragraph{MHA special case.}
For multi-head attention ($G=1$) the group sum collapses to a single term $c\,\tau/(1-\tau)$,
with $\tau=\sum_{b\in S^c}P_b$ the total dropped mass. The positive constant $c$ does not affect the ordering in this case; moreover, the penalty is monotonically increasing in $\tau$, and $\tau$ is additive over the dropped blocks. Therefore, minimizing the bound is exactly keeping the top-$k$ blocks by mass $\mass$, equivalently by log-mass $\ln\mass$,
so \Cref{ass:gamma,ass:linear} are unnecessary.

\section{Cumulants of a block's key distribution}
\label{app:cumulants}

We justify~\eqref{eq:cgf-cumulants}. The cumulants are the derivatives of $\KX(q)=\ln\MX(q)$ at
$q=0$; the standard cumulant--moment relations give the first two as the mean and covariance of $X$,
\begin{equation}
  \kappa_1=\E_X[X],\qquad
  \kappa_2=\E_X\!\big[X X\T\big]-\E_X[X]\,\E_X[X]\T
    =\E_X\!\big[(X-\E_X[X])(X-\E_X[X])\T\big],
\end{equation}
so $\kappa_2$ is the second central moment. Since $X$ is
uniform over the block's $L$ keys $\{k_r\}_{r\in b}$, these evaluate to
$\kappa_1=\tfrac1L\sum_{r\in b}k_r=\kb$ and
$\kappa_2=\tfrac1L\sum_{r\in b}(k_r-\kb)(k_r-\kb)\T=\Sigb$.

\section{RULER-style supervised fine-tuning and task selection}
\label{app:sft}

\paragraph{RULER-style SFT.} Long-context retrieval is difficult to elicit from small models by
pretraining alone. To obtain a meaningful long-context retrieval signal at a scale we can iterate on,
we perform supervised fine-tuning (SFT) on generated RULER-style long-context data~\cite{ruler} for
our $\approx$1B-parameter models. This lets $\approx$1B models reach accuracies where the differences
between selection mechanisms are measurable, so we can compare attention mechanisms and architecture
changes at a small, fast-to-train scale rather than at frontier-scale pretraining. The SFT data uses
the same task templates as RULER but is disjoint from the evaluation instances (\Cref{sec:experiments}).

\paragraph{Omitting the QA tasks.} RULER includes reading-comprehension question-answering (QA) tasks,
which we omit from our 11-task configuration for two reasons. First, we do not SFT on any QA-style
data, so these tasks are out of distribution for our fine-tuned models. Second, QA accuracy is limited
by the world knowledge and comprehension acquired during pretraining, which our models largely lack
after pretraining on LongCrawl64~\cite{longcrawl64}: even full (dense) attention scores at most
$\approx$10\% on the QA tasks. Because every variant sits near this floor, the QA tasks contribute noise
rather than signal about selection quality, so we exclude them and report the mean over the remaining
11 tasks.

\end{document}